\documentclass[runningheads]{llncs}
\usepackage[T1]{fontenc}
\usepackage{graphicx}
\usepackage{latexsym}
\usepackage{amssymb}
\usepackage{amsmath}
\usepackage{booktabs}
\usepackage{enumitem}
\usepackage{color}
\usepackage{lipsum}

\usepackage{comment}
\usepackage{arydshln}

\usepackage{tikz}
\usepackage{mdframed}
\usepackage{algorithm}
\usepackage{algpseudocode}

\begin{document}
\title{An Epistemic Human-Aware Task Planner which Anticipates Human Beliefs and Decisions\thanks{\footnotesize Accepted for publication in the proceedings of the International Conference on Social Robotics (ICSR) 2024.
}}
%
%
\author{Shashank Shekhar\inst{1} \and
Anthony Favier\inst{1,2} \and
Rachid Alami\inst{1,2}}
\authorrunning{S. Shekhar et al.}
%
\institute{LAAS-CNRS, Université de Toulouse, CNRS, INSA, UPS, Toulouse, France \and
Artificial and Natural Intelligence Toulouse Institute (ANITI) 
\\
\email{\{shashank.shekhar,afavier,rachid.alami\}@laas.fr}
}
\maketitle              
\begin{abstract}
We present a substantial extension of our Human-Aware Task Planning framework, tailored for scenarios with intermittent shared execution experiences and significant belief divergence between humans and robots, particularly due to the uncontrollable nature of humans.
Our objective is to build a robot policy that accounts for uncontrollable human behaviors, thus enabling the anticipation of possible advancements achieved by the robot when the execution is not shared, e.g., when humans are briefly absent from the shared environment to complete a subtask. 
But, this anticipation is considered from the perspective of humans who have access to an \textit{estimated} robot's model. 
To this end, we propose a novel planning framework and build a solver based on AND/OR search, which integrates knowledge reasoning, including situation assessment by perspective taking.
Our approach dynamically models and manages the expansion and contraction of potential advances while precisely keeping track of when (and when not) agents share the task execution experience.
%
The planner systematically {\em assesses} the situation and ignores worlds that it has reason to think are impossible for humans. 
%
Overall, our new solver can estimate the distinct beliefs of the human and the robot along potential courses of action, enabling the synthesis of plans where the robot selects the right moment for \textit{communication}, i.e. informing, or replying to an inquiry, or defers ontic actions until the execution experiences can be shared. 
Preliminary experiments in two domains --- one novel and one adapted --- demonstrate the framework's effectiveness.

%



\end{abstract}
\section{Introduction}


Studies in psychology and cognitive science within the domain of joint actions suggest that humans consider each other's actions and beliefs, indicating that they model each other's tasks when planning~\cite{schmitz2017co,kourtis2014attention,sebanz2009prediction}. 
Therefore, it is important if not key for success to be able to estimate or anticipate situations of divergence in beliefs and how that can be detrimental to collaborative activities.

In joint action scenarios, where partners work toward a shared goal, individuals often form expectations of their partner’s actions based on their own mental models, which may be flawed. 
When separated, they rely on these models to estimate their partner’s progress, but inaccuracies or incomplete understanding of their collaborator's beliefs and capabilities can lead to misaligned expectations.
This cognitive bias shows the complexities of effective collaboration between a robot and a human on a shared task. 
And, highlights the need for robust frameworks to manage these differences.

We take the first step in this paper towards building a planning framework for human-robot collaboration that generates robot policies to address the issue of inaccurate mental models. 
Our proposed strategy integrates by adapting tools developed for epistemic planning~\cite{BolanderA11}, Dynamic Epistemic Logic (DEL)~\cite{KR2021-12}, and human-aware planning~\cite{alili2009planning,CirilloKS09,buisan:hal-03684211,FavierSA23,UnhelkarLS20}.

\begin{figure}[!t]
   \centering   \includegraphics[width=1\textwidth]{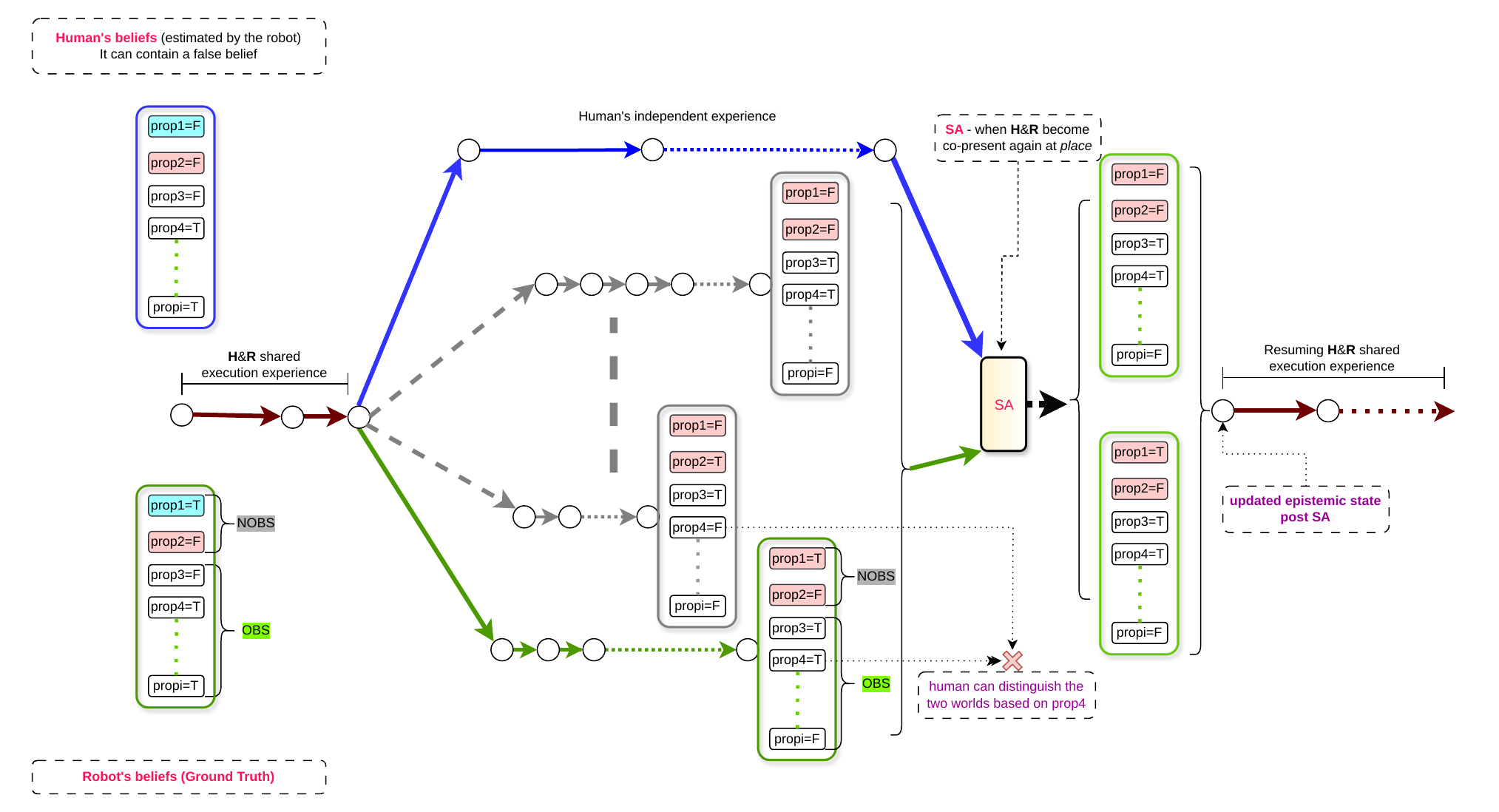}
   \caption{ \em Our planning framework is endowed with the ability to make the difference between \textbf{H}\&\textbf{R} shared and individual execution experiences in the planned activities. It can anticipate potential belief divergence between \textbf{H}\&\textbf{R} and also estimate the updated beliefs of \textbf{H} when they meet again {\em(situation assessment (SA))} based on a distinction between observable and non-observable facts. This will be used to plan communicative actions or adapt the \textbf{R}'s plan to ensure the shared experience of some actions. 
   In this diagram, we roughly depict what happens when \textbf{H}\&\textbf{R} no longer share the execution experience, \textbf{H} has independent 
   experience (\textbf{blue}), while \textbf{R} progresses towards the goal (\textbf{green}), with anticipated traces 
   (in \textbf{gray}) depicting other estimated courses of action that the robot can choose along with the green trace but from the \textbf{H}'s perspective. 
   Upon co-presence at \textit{place}, \textit{SA} eliminates impossible worlds, e.g., those with state property \textit{prop4=F} (since it is \textit{observable}), aiding \textbf{H} to ignore wrongly estimated worlds. 
   }   
   \label{fig:teaser_figure}
   \vspace{-0.5cm}
\end{figure}

To this end, we propose a novel \textit{epistemic human-aware task planning} framework. It substantially extends our past works and enables the robot to estimate, anticipate, and adapt to scenarios in which an uncontrollable human and a robot have disrupted shared execution experiences. 
Specifically, it considers the human's perspective and estimation regarding the potential advances achieved by the robot, even when the exact progress is not directly experienced by the humans, who may hold an \textit{incorrect} robot model.


\begin{sloppypar}
In addition, we build an AND/OR search-based offline planner that facilitates Theory of Mind (ToM) by integrating knowledge reasoning and incorporating situation assessment. 
It dynamically manages the evolution or contraction of estimated possible worlds from the human's point of view.
This helps the planner to prepare itself with a set of worlds that humans would consider possible.
\end{sloppypar}

Our framework adapts tools developed in the literature, including those for DEL-based epistemic planning. 
However, as we will soon show, also a minor contribution, is that it offers more flexibility. Unlike the majority of epistemic planners, our framework does not require scripting all the effects on the beliefs of every agent in the action models as input.

\begin{figure}[t!]
    \centering        \includegraphics[width=0.5\textwidth]{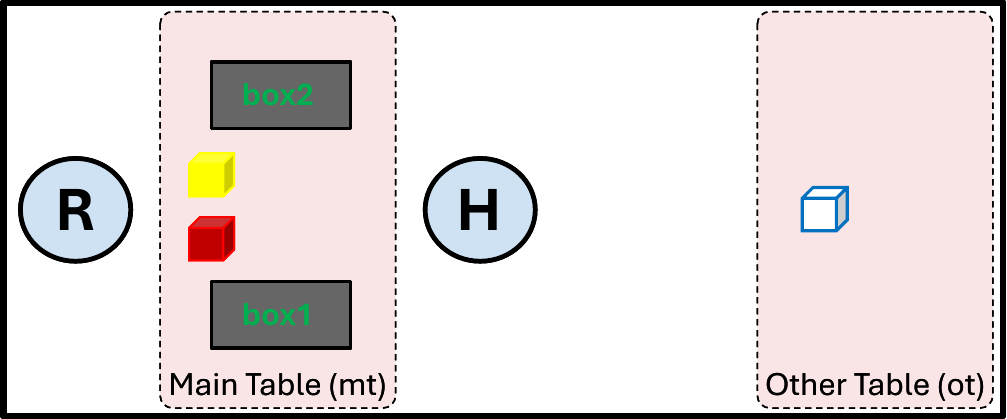}
    \caption{ \em
    Three cubes $c_r$ (red), $c_y$ (yellow), and $c_w$ (white) are shown. $c_r$ and $c_y$ are placed on \textit{mt} (main table), and $c_w$ is on \textit{ot} (other table). There are two boxes, $box_1$ and $box_2$, placed on \textit{mt}, which can be either transparent or opaque. The shared task is to organize the cubes in a way that cubes from one table are placed in one box. The choice of which box is flexible as long as each table's cubes end up in separate boxes.}   
    \label{fig:cube_org_scenario}
    \vspace{-0.5cm}
\end{figure}

Figure~\ref{fig:teaser_figure} provides a rough illustration of a single plan trace, showing what happens when agents share execution experiences and when they do not in the process of achieving the shared task.

Thanks to our novel framework and the planner, it enables the robot to take proactive steps, such as anticipating humans to be \textit{inquiring} about an unknown variable's value, \textit{communicating} relevant information without being annoying (\textit{e.g.}, not verbalizing a fact already known to them), or deferring \textit{executing} an action until \textbf{H}\&\textbf{R} reunite, thus reducing the ambiguities for \textbf{H}.

We outline our \textit{key contributions} specific to this paper as follows, directly addressing the primary problem discussed above:
\begin{itemize}[label=$\bullet$]
    \item We have introduced \textit{human mental model} in our previous framework. 
    \item We developed a novel, sound planning algorithm that integrates human situation assessment and anticipates the inferences humans will make upon observing the new world state.
    \item Non-controllability is not unique here, but we extend~\cite{buisan:hal-03684211,FavierSA23,ShekharFAC23} to address events of (non-)shared execution experiences and managing beliefs. 
    We present enriched models for co-presence, observability, and situation assessment.
    \item We show our planner's effectiveness with experimental results in two domains, one is novel (our case study) and another one is adapted from~\cite{FavierSA23}.
\end{itemize}

The paper is structured as follows. A case study is presented, followed by background information on necessary tools. Next, we describe our proposed framework, followed by the AND/OR search-based algorithm. The subsequent section discusses related work, followed by preliminary experiments showing the effectiveness of the framework in diverse scenarios. Finally, we conclude our work.


\section{The Cube Organization Case Study}

Figure~\ref{fig:cube_org_scenario}  illustrates the task of organizing cubes into boxes. 
The shared \textbf{HR} task requires that cubes from different tables be placed into separate boxes.


Say only \textbf{H} is capable of moving around and exhibits unpredictable behavior (\textit{nondeterminism}), such as moving to the other table (\textit{ot}) to retrieve cubes, while \textbf{R} may continue to act. 
From the \textbf{H}'s perspective, \textbf{R} may move some or all of the cubes from the main table (\textit{mt}) and place them into one of the boxes, or it may choose to take no action at all. 
Upon returning to the main table \textit{mt}, \textbf{H} may discover that some, none, or all of the cubes originally on \textit{mt} are missing, indicating that they have been placed in one of the boxes.

If \textbf{R} places some cubes from \textit{mt} into one of the boxes, \textbf{H} will only learn about this decision upon encountering transparent boxes. 
But when opaque, \textbf{R} has several options: it can communicate, wait for \textbf{H} to inquire, or select a remaining cube of \textit{mt} to place in the correct box when \textbf{H} and \textbf{R} are co-present.


Planning is done from the robot's perspective, taking into account \textbf{R}'s and \textbf{H}'s task models.
The human collaborator has an approximation of the robot's model, which enables them to anticipate the robot's action. 
We later provide more details on these models and about their accuracy and falsity. 

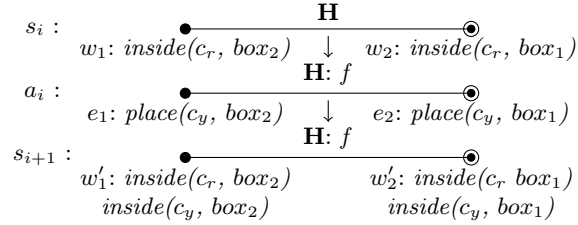
\begin{figure}[t!]
\vspace{-10pt}
\centering
\begin{tikzpicture}[scale=0.95]
\filldraw[fill=black] (0,0) circle (2pt) node[below] {$w_1$:~\textit{inside($c_r$, $box_2$)}};
\filldraw[fill=white] (4,0) circle (3pt) node[below] {$w_2$:~\textit{inside($c_r$, $box_1$)}};
\filldraw[fill=black] (4,0) circle (1.5pt);
\draw (0,0) -- node[above] {\textbf{H}} (4,0);
\node[anchor=center] at (-2,0) {$s_i$~:~};

\filldraw[fill=black] (0,-0.9) circle (2pt) node[below] {$e_1$:~\textit{place($c_y$, $box_2$)}};
\filldraw[fill=white] (4,-0.9) circle (3pt) node[below] {$e_2$:~\textit{place($c_y$, $box_1$)}};
\filldraw[fill=black] (4,-0.9) circle (1.5pt);
\draw (0,-0.9) -- node[above] {\textbf{H}:\textit{~f}} (4,-0.9);
\draw[->] (2,-0.1) -- node[right] {~} (2,-0.4);
\node[anchor=center] at (-2,-0.9) {$a_i$~:~};

\draw[->] (2,-1.0) -- node[right] {~} (2,-1.3);
\filldraw[fill=black,text width=3cm, align=center] (0,-1.8) circle (2pt) node[below] {$w'_1$:~\textit{inside($c_r$, $box_2$) inside($c_y$, $box_2$)}};
\filldraw[fill=white,text width=3cm, align=center] (4,-1.8) circle (3pt) node[below] {$w'_2$:~\textit{inside($c_r$ $box_1$) inside($c_y$, $box_1$)}};
\filldraw[fill=black] (4,-1.8) circle (1.5pt);
\draw (0,-1.8) -- node[above] {\textbf{H}:\textit{~f}} (4,-1.8);
\node[anchor=center] at (-2,-1.8) {$s_{i+1}$~:~};
\end{tikzpicture}
\vspace{-10pt}
\caption{ \em
We represent a state $(s_i)$, action $(a_i)$, and how applying $a_i$ in $s_i$ leads to next state $(s_{i+1} = s_{i} \otimes a_{i})$. $f$ is a formula that captures if \textbf{H}\&\textbf{R} were co-present when the events took place. 
Common facts for both worlds, such as \textit{opaque($box_1$)}, are not shown. Also, each world is fully defined, with either an atom or its negation holding true.}
\label{fig:consolidated-es-es-nes}
\vspace{-0.4cm}
\end{figure}



\section{Background}
\subsubsection{Dynamic Epistemic Logic (DEL).}
%
We focus on epistemic languages $(\mathcal{L_K})$, a state $(s$ --- comprising a set of worlds $w_i)$, an action $(a$ --- comprising a set of events $e_i)$, and state transitions (via the \textit{cross product} $\otimes$ operator) as derived from the literature~\cite{BolanderA11,KR2021-12}, with necessary simple adjustments for our needs. 
For other basic concepts like \textit{indistinguishability} and \textit{equivalence relation}, \textit{perspective shift}, and \textit{truth} of epistemic formulas, readers are referred to the cited literature.


Here, we focus on the essential DEL concepts necessary to build the framework, using examples from use case study. Recall the requirements for the task.

\begin{sloppypar}
    \begin{example}
    Say the task is in the state $s_i$ (Fig.~\ref{fig:consolidated-es-es-nes}), in which $c_r$ is inside $box_1$ and both the boxes are opaque, and the robot holding $c_y$ and the human comes back with $c_w$, and assesses the situation. 
    We assume that the human can see the robot holding $c_y$. The epistemic state $s_i$ 
    such that $s_i \models K_{\textbf{R}}inside(c_r, box_1)$, but concerning the human partner, $s_i \models \neg K_{\textbf{H}}inside(c_r, box_1) \wedge \neg K_{\textbf{H}}inside(c_r, box_2)$. Here, $K_ip$ represents agent $i$ knows that the literal $p$ is true.
\end{example}
\end{sloppypar}

\begin{sloppypar}    
\begin{example}
The next state $s_{i+1}$ is such that, the epistemic action the robot will execute in epistemic state $s_i$ is $a_i$ that is placing $c_y$ in the {\em correct} box. 
We describe how the next epistemic state $s_{i+1}$ looks like \textit{when} and \textit{when not} \textbf{H}\&\textbf{R} are \textit{co-present} (i.e., whether they share this experience) during execution:
%
An indistinguishability relation is only for \textbf{H} when the formula {\em $f$}, e.g., {\em at(\textbf{R}, place) \& not(at(\textbf{H}, place))}, holds. 
\textbf{R} always knows that the designated world is $w_2$. That means if the human is co-present, they will know that the real world is $w_2$.
\end{example}
\end{sloppypar}

\subsubsection{Human-Aware Task Planning.}
We briefly discuss the human-aware task planning paradigm here. HATP/EHDA~\cite{buisan:hal-03684211} comprises a dual Hierarchical Task Network (HTN) based task specification model. It is a recently proposed planner that \textbf{e}stimates and emulates \textbf{h}uman \textbf{d}ecisions and \textbf{a}ctions for HRC.
It solves problems in a turn-taking fashion, as formalized in our previous work~\cite{FavierSA23,FavierSA22a}. The following language adheres to this framework for easier understanding.

Consider the \textit{human-aware task planning problem}, $\mathcal{P}_{rh}$ and \textit{implicitly coordinated joint solution} defined (Definitions 5 \& 6, respectively) in~\cite{FavierSA22a}. 

\textbf{R} and \textbf{H} have their action models, beliefs ($Bel(.)$), agenda or task networks ($tn$), plans, and more, collectively comprising $\mathcal{P}_{rh} = \langle \mathcal{M}_R, \mathcal{M}_H \rangle$.
More specifically, \textbf{R} 
has its estimated beliefs, $s_0^r$. We consider it as the \textit{knowledge} for \textit{``ground truth''} in the planner's reference, versus what \textbf{R} estimates to be believed by \textbf{H}, $s_0^h$, by perspective taking. 
$s_0^h$ may include a literal that is not true (\textit{false belief} -- e.g., \textit{prop1} in Fig.~\ref{fig:teaser_figure}) from \textbf{R}'s perspective and can be \textit{corrected}.
%

%

We extended HATP/EHDA in~\cite{FavierSA23}, which adeptly anticipates human false beliefs for better collaboration based on (non-) shared execution experience.

To achieve that, \textit{situation assessment} processes based on co-presence are integrated into the planning framework of HATP/EHDA.
This enhances the planner to be pertinent to capturing what humans can observe and infer in their surroundings. 
It assesses the detrimental effects of humans’ incorrect beliefs on the task at hand. 
As a result, \textbf{R} plans to communicate minimally and proactively.
%

We demonstrated in our previous work how to handle false beliefs (of first order) and situate the research broadly within the literature. In this paper, we extend and model knowledge up to level two, enabling us to handle \textbf{HR} collaboration more realistically and allowing us to incorporate communication in a more practical way. 
We detail all these aspects as we proceed. 


\section{The EHATP Planning Framework}
%


We consider that the human maintains an estimated model for the robot $\mathcal{M}^R_H$, which can be \textit{incorrect} compared to $\mathcal{M}_R$. 

The epistemic HATP (EHATP) framework considers three models: $\mathcal{M}_R$, $\mathcal{M}_H$ and $\mathcal{M}^R_H$. 
While $\mathcal{M}_R$ guides the planning of \textbf{R}'s actions and $\mathcal{M}_H$ helps estimate and emulate \textbf{H}'s decisions and actions. But, using $\mathcal{M}^R_H$, \textbf{H} \textit{``expects''} and \textit{``predicts''} certain robot behavior (from their own perspective) both, respectively, when they are co-present and when they are not. Note that, each model has their own dedicated components like $Bel(.)$ and $tn$ as defined earlier.

The majority of the models' components remain static, but for each model, its task network ($tn_{\phi}$) and belief ($Bel(\phi)$) components are dynamic, where $\phi$ denotes an agent (or agent perspective). 
Except for belief, we assume that components like the robot's action model and task network are accurately estimated by \textbf{H}. 
This allows us to focus on the key aspects relevant to this paper. For other incorrectly estimated components of $\mathcal{M}^R_H$, we suspect a possible generalization utilizing concepts developed in~\cite{SreedharanCK21} and intend to explore this in the future.


\subsection*{Planning Workflow}
\begin{sloppypar}
We focus on only the dynamic parts. 
The initial epistemic state $s_0$ (with the only world to begin with and that is also the designated world $w_d$) is provided as an input. 
In general, each world $w_j$ in an epistemic state $s_i$ represents $\langle (Bel(R),tn_r), (Bel(H), tn_h), (Bel(R_H), tn_{r_h})\rangle$. 
It also includes the only designated world $w_d$ always known to \textbf{R}. 
Note that these worlds are indistinguishable for \textbf{H}, but human knows that the robot can always distinguish them and that the robot can identify $w_d$. 
Also, the human knows that, if $w_j$ is the designated world, then $Bel^{ij}(R_H)$, is the reality as they do not have access to the facts appearing in $Bel^{ij}(R)$.
Here, we consider that $Bel(H)$ is equal to $Bel(R_H)$, but they can be different from $Bel(R)$ and can contain false (human) beliefs.
\end{sloppypar}

The robot, an epistemic state $s_i$ and possible worlds $w_{j}$ in it are considered. 
We compute the set of all possible primitive actions, computed by all feasible decompositions, based on 
$(Bel(R),tn_r)_{ij}$, and whether it is different than the set of primitive actions based on the allowed decompositions w.r.t. $(Bel(R_H), tn_{r_h})_{ij}$. 
The idea is to align these decompositions, w.r.t. each $w_j$, in a way that the human can correctly estimate the progress the robot may achieve, thus utilizing the human's capacity for anticipating.  
If there is a difference, we identify the \textit{relevant} facts in $Bel^{ij}(R)$ that need to be corrected in $Bel^{ij}(R_H)$, to align the decompositions. 
To achieve that, we adapt our earlier approach presented in~\cite{FavierSA23}. That is, one can plan minimal communication, possible to schedule ahead of time during offline planning when communication is allowed.
Eventually, communication will also fix $Bel^{ij}(H)$, accordingly. 
However, $Bel^{ij}(H)$ and $Bel^{ij}(R_H)$ can still have \textit{non-relevant} false beliefs compared to the ground truth ($Bel^{ij}(R)$).


Next, the planner computes the \textbf{R}'s next real action based on its task network $tn^{id}_R$ in the designated world $w_d$ of $s_i$, we call it the \textit{designated} event.  
It also computes other non-designated events based on respective decompositions in each world $w_{j}$ of $s_i$. 
(An event and a possible real action including \textit{noop}s are used interchangeably.) 
In other words, the planner computes a set of all possible decompositions based on what \textbf{H} can anticipate, that means by taking into account each $(Bel(R_H), tn_{r_h})_{ij}$. 
These are all the anticipated events that can happen due to the robot acting, but the designated event may or may not be assessed depending on \textit{co-presence}.
%
All the decompositions (\textit{i.e.}, the set of the first primitive action in each refinement) together form an epistemic action $a_i$.



    

\paragraph{Executing an Epistemic Action in a State:}
Based on the cross-product operation ($\otimes$), the state transition is computed as $s_{i+1} = s_i \otimes a_i$. 
In our planning algorithm (Algorithm~\ref{algo:and-or-search}, Line~8), we model the scenario as follows: if \textbf{H}\&\textbf{R} are \textit{co-present}, then \textbf{H} can distinguish between the actual event (the real action performed by \textbf{R}) and other estimated events. 
Otherwise, \textbf{H} perceives each event as a possible action by \textbf{R}.
When co-present, \textbf{H} assesses the execution of \textbf{R}'s real action, thus narrowing down the possibilities over $w'_{j}$'s in $s_{i+1}$ --- captured by $\otimes$ (\textit{ref~Fig.~\ref{fig:consolidated-es-es-nes}}).




Within each world of the new epistemic state, belief components, i.e., $Bel(R)$, $Bel(H)$, and $Bel(R_H)$ are updated corresponding to the possible robot action (either \textit{real} or \textit{anticipated}) that is a part of epistemic action $a_i$. 
Also, the task networks concerning $\mathcal{M}^R$ and $\mathcal{M}^R_H$ are updated in each world, accordingly.

        
\paragraph{When The Human Acts:}
\textbf{H} acts only if their next real action, w.r.t. a possible decomposition, is applicable in all possible worlds. 
I.e., for each $w_j$ in $s_{i+1}$, applicability of the action is examined in every $(Bel(H), tn_h)_{i+1,j}$. 
\textit{Two key issues} at this stage are: 
First, humans can act based on a false belief (if consistent throughout all the worlds), or a true belief w.r.t. the ground truth in every $w_j$. 
We handle false belief scenarios the way it is addressed in the literature, that is, by finding out relevant belief divergence and handling it via communication~\cite{FavierSA23}. 



Second, we also know that a boolean variable, $p$, that \textbf{H} is uncertain about at this stage, which holds only in some worlds, is due to disrupted shared execution experiences.
If $p$ is a precondition of the task refinement process, then \textbf{H} can initiate communication, or \textbf{R} can inform \textbf{H} about $p$. And, if co-present, \textbf{R} can also act to implicitly share $p$'s value such that there is some correlation between that action and $p$. 
Here, we focus on explicit communication, while sharing $p$'s value by changing the environment is left for the future. 

\subsubsection{Handling \textbf{H}\&\textbf{R} Communication.}
We introduce two types of actions and they become a part of the deliberation process. 
First, \textit{ask-$p$} -- human inquires about $p$ from \textbf{R}, and, second \textit{inform-$p$} -- \textbf{R} informs them of the status of $p$. 


At this stage, we create two specialized versions of state $s_{i+1}$: one prioritizing human inquiries, \textit{ask-$p$}, and the other prioritizing robot updates, \textit{inform-$p$}. 
Communication tasks are adjusted into respective networks appropriately.

\begin{sloppypar}
\vspace{-0.4cm}
\begin{algorithm}[t]
\caption{{\em AND/OR Planner using Breadth-First Search. 
}}
\begin{algorithmic}[1]
\State \textbf{Input:} \textit{A HAETP task}
\State \textbf{Output:} \textit{A joint solution} or {\em failure}
\State $root\_epi\_state \gets \langle \mathcal{M}, w_d \rangle $
\Comment{(focusing just on the dynamic parts) each world in $w \in W$ contains
$(\langle s^r_0, tn_{r,0}\rangle, \langle s^h_0, tn_{h,0}\rangle, \langle s^{r_h}_0, tn_{r_h,0}\rangle)$ } and $W = \{w_d\}$
\State $queue.\text{enqueue}(root\_epi\_state)$

\While{$queue$ is not empty}
    \State $curr\_node' \gets \text{queue.dequeue()}$
    \State $curr\_node \gets \textcolor{red}{\textit{Situation Assessment($curr\_node'$)}}$
    \State $successors \gets \textcolor{red}{\textit{Expand($curr\_node$)}}$
    \If{$successors$ $\neq$ $\emptyset$} 
        \For{$\text{successor}$ \textbf{in} $\text{successors}$}        
        \State $queue.\text{enqueue}(successor)$
        \EndFor
    \Else
        \State \textit{eval}$(curr\_node)$ \Comment{assign it \textit{DONE} or \textit{DEAD}} 
        
        \State $\textit{propagate\_revised\_status}(curr\_node)$
    \EndIf
    \If{$\textit{root\_solved}(root\_epi\_state)$}
        \State \Return $\textit{extract\_joint\_solution()}$
    \EndIf   
\EndWhile
\State \Return \textit{failure}
\end{algorithmic}
\label{algo:and-or-search}
\end{algorithm}
\end{sloppypar}

\subsubsection{Situation Assessment.}

Assessing the status of a state property depends on a broader context, which determines whether it can be observed or only inferred by attending the action execution affecting it. 
Knowledge rules were used to address this aspect~\cite{ShekharFAC23}. 
For example, \textbf{H} can view the current status of the variable \textit{inside$(c_r, box_1)$} as \textit{true} if they meet the requirements of the rule's antecedent formula, e.g., being at the main table, $box_1$ is transparent, and $c_r$ is inside $box_1$. 
While formally defined below, we depict how the \textit{SA} process works in Figure~\ref{fig:teaser_figure}. 

\begin{definition}
The {\em situational assessment} (SA) process considers our {\em observation} process and a state $s_i$, producing an updated epistemic state $s'_i$. 
This iterates over each world $w_j$ in $s_i$, removing it if human can distinguish it from $w_d$.
\label{def:sa_epip_state}
\end{definition}

\begin{sloppypar}
%
%
\end{sloppypar}

\section{AND/OR Search based EHATP Planner}
Algorithm~\ref{algo:and-or-search} takes the EHATP problem as input, producing an output as either a \textit{failure} or an optimal worst case joint solution. 
It is an implementation of the classic AND/OR search using rooted graphs. 
When the \textit{root} node is \textit{DONE}, the joint solution policy is extracted (\textit{extract\_joint\_solution()}), in Lines~17 \& 18.

We consider the root 
node ($root\_epi\_state$) and the subsequent actor, either \textbf{R} or \textbf{H}, to begin the plan exploration (Line~3).
Within the loop, in Line~6, we select a node/state from $queue$, and next call the \textbf{\textit{Situation Assessment(~)}} subroutine.  
At this stage, the planner already knows whether agents were co-present and whether \textbf{H} assessed the designated event.
It ignores the worlds distinguishable from the designated world (\textit{Definition~\ref{def:sa_epip_state}}). 
The scenario where a human transitions to the \textbf{R}'s location and subsequently becomes co-present is particularly interesting.
Another significant subroutine, \textbf{\textit{Expand(~)}}, previously discussed in the EHATP framework's planning workflow, is invoked in Line~8.
The children created after \textbf{R} expands the popped node are \textit{AND} nodes. Conversely, when \textbf{H} expands the popped node, \textit{OR} nodes are created. 


In Line~14, we evaluate the current node.
If both $tn_r$ and $tn_h$ are fully decomposed in the designated world of $s_i$, we execute an auxiliary action with a precondition that the task network is fully decomposed. If both agents can execute it individually, it signifies that agents believe that the shared task has been achieved.
In Line~15, it propagates the status of this node to its immediate parent, which then further propagates the status upwards.



\paragraph{\em \textbf{The Post-processing Step.}}
Post-processing of the joint solution is done based on whether \textbf{H}\&\textbf{R} are co-present. 
When \textit{co-present}, we follow a turn-taking approach, but when \textit{not} co-present, their actions are parallelized. 
This involves executing the AND/OR policy, and identifying where \textbf{H}\&\textbf{R} separate and reunite. 
We then group the agents' actions in between to form pairs. 

\begin{sloppypar}
\paragraph{\em \textbf{Runtime Analysis of Reasoning in EHATP.}}
In the worst-case scenario, roughly, the runtime is influenced by the robot’s available choices ($m$) in the absence of the human at each step, as these choices are crucial for updating the human mental model ($\mathcal{M}^{R}_{H}$) correctly. 
This is then multiplied by the number of choices ($b$) the human has to progress with the task when they are copresent.

We introduce a parameter $\mathit{K}$, which represents the maximum \textit{\#actions} \textbf{R} can perform when \textbf{H}\&\textbf{R} are not copresent. So, the runtime complexity can be $O(b \times m^{K})$ from the point they separate and reunite again, in terms of epistemic state exploration \textit{s.t.} the maximum number of possible worlds in a state is $m$. 
We assume $\mathcal{M}^{R}_{H}$ and $\mathcal{M}_{R}$ are aligned at this stage when they separate.
%
%
\end{sloppypar}


\section{Related Work}

\begin{sloppypar}
\noindent \textbf{Human Robot Collaboration (HRC):}
Generating the robot's behavior while considering the existence of humans, known as human-aware planning and decision-making~\cite{CirilloKS09,alili2009planning,UnhelkarLS20,lallement2014hatp,lemaignan-2017,darvish2020hierarchical,CramerKD21}. 
Also, it can do reasoning for task allocation~\cite{roncone2017transparent,ramachandruni2023uhtp}. 
Communication is an essential key to successful HRC, which is used to align an agent's belief, clarify its decision or action, fix errors, etc.~\cite{tellex2014asking,lemaignan-2017}.
We extend this research line but have not found studies addressing human anticipation and divergent beliefs in disrupted execution experiences. 
\end{sloppypar}

\noindent \textbf{Models, Planning Approaches, and Solutions:}
Several planning models are applied in the context of HRC planning, including HTNs~\cite{lallement2018hatp,roncone2017transparent,cheng2021human}, POMDPs~\cite{UnhelkarLS19,roncone2017transparent,UnhelkarLS20}, AND/OR graphs~\cite{DarvishSMC21}, etc.
HTNs use both abstract and non-abstract tasks to form hierarchical networks, while AND/OR graphs cover causal links among subtasks and 
depth-first search is used in planning~\cite{GombolayJSSS16}.



\noindent \textbf{Epistemic Planning:}
The epistemic planning framework, in~\cite{KR2021-12}, holds promise for capturing key elements of ToM in autonomous robots. For HRC, the framework lays the groundwork for implicit coordination through perspective shifts~\cite{engesser2017cooperative}. 
By adapting this framework and focusing on the robot's perspective, it may serve as a basis for addressing the core problem we have aimed at with the shared mental model~\cite{nikolaidis2012human}, albeit without assuming imperfectly estimated model ($\mathcal{M}_H^R$).


\noindent
\textbf{Explainable AI Planning (XAIP):}
In general, XAIP focuses on human-aware systems providing explanations of their behavior~\cite{KambhampatiSVZG22}. E.g., a system might explain the correctness of its plan and the reasoning behind its decision based on its own model.
The model reconciliation approach~\cite{SreedharanCK21}, assumes that the human possesses a disparate model of the robot's behavior ($\mathcal{M}^R_H$ instead of $\mathcal{M}_R$). 
It avoids unnecessary explanations by identifying the specific differences between the two models and only generates explanations where needed. 
Essentially, it suggests changes to $\mathcal{M}^R_H$ to optimize the robot's plan based on that revised $\mathcal{M}^R_H$.
The approach calculates the optimal explanations by identifying relevant discrepancies and communicating only the necessary information to align the models.
We suspect a possible generalization of our approach while adapting this method to ``correct'' only what is necessary to align decompositions.

\begin{table}[!t]
    \centering
    \resizebox{0.65\linewidth}{!}{
    \begin{tabular}{c|c|c|c| c|c|c}
        \hline
         \textit{inst} & $\mathit{K}$ & \textit{comm} & \textit{\#states} & $|W|$ &\textit{\#leaves} & \textit{time (ms) $\times$ $10^5$}  
         \\ \hline
         P1 (2,2,T) & 2 & N & 218 & 4 & 3 & 0.089  
         \\ 
         P2 (2,2,O) & 2 & Y & 236 & 4 & 3 & 0.141  
         \\ 
         P3 (3,2,T) & 2 & N & 1643 & 7 & 6 & 5.906
         \\ 
         P4 (3,2,O) & 2 & Y & 2003 & 7 & 6 & 9.816
         \\ 
         P5 (3,2,T) & 4 & N & 4107 & 14 & 5 & 99.81 
         \\ 
         P6 (3,2,O) & 4 & Y & 5607 & 14& 5 & 125.3 
         \\ \hdashline
        Cooking~1 & 2 & Y & 603 & 3 & 5 & 0.382
        \\
        Cooking~2 & 3 & Y & 1054 & 4 & 5 & 1.474 
        \\Cooking~3 & 4 & Y & 1800 & 5 & 5 & 5.301
         \\\hline
    \end{tabular}
    }    
    \caption{\em The planner's performance metrics are presented. {\em inst} describes the instance number; for the first domain, this includes the number of cubes and boxes, as well as the property of the boxes (T for transparent, O for opaque). {\em comm} indicates whether communication is used. The reported metrics include the total number of explored states ({\em \#states}), the worst-case number of worlds evaluated in a state ($|W|$), the number of traces in the final AND/OR solution tree ({\em \#leaves}), and the execution time (measured in $10^5$ ms). Two sections of the table include instances from the respective domains.}
    \label{tab:quant_res_1}
    \vspace{-0.7cm}
\end{table}

\section{Empirical Evaluation}
%
%
We implemented our planning system using Algorithm~\ref{algo:and-or-search} in Python. 
It is based on the latest version of HATP/EHDA code~\cite{buisan:hal-03684211}.


No standard planners are available for comparison to our knowledge. 
We will gauge the performance of our planner against the one from~\cite{FavierSA23}, which provides limited support for scenarios with disrupted shared execution experience. 
It is worth noting that directly comparing their runtime would not be entirely fair, as our planner operates with a richer representation.

\noindent
\textbf{Domain Description:} We test the planner in our use case domain and the cooking domain adapted from~\cite{FavierSA23}, on a variety of problems.

In the adapted scenario, both \textbf{H} and \textbf{R} are tasked with preparing dinner. The main activities involve \textit{cutting} (R), \textit{washing} (R) vegetables, \textit{putting} (R) them on the stove with a pan and \textit{seasoning} (R) them. Depending on the vegetables, seasoning can occur before or after they are placed in the pan, but always after washing. \textbf{H} is responsible for \textit{bringing} (H) spices and other ingredients from the pantry and \textit{mixing} (H) them in the pan, but only after the vegetables have been boiled (i.e., the effect of the \textit{putting} action). \textit{Serving} (H) dinner can only happen after the spices and seasoning have been mixed. Actors appear in \texttt{()}. 
Effects of washing and seasoning are non-observable.

The decision to bring ingredients separates \textbf{H} from \textbf{R}. Despite this adaptation, \textbf{H} can still choose when to leave the kitchen for the pantry.

\subsection{Experiments}

\paragraph{\em \textbf{Analyzing the Impact of $\mathit{K}$ and Non-Determinism.}}
Algorithm~\ref{algo:and-or-search} highlights a rapid growth in the size of the epistemic state in terms of the number of worlds which directly correlates with $\mathit{K}$ that is the maximum $\#actions$ the robot can perform when the experience is not shared. 
The sequencing of actions significantly influences the range of potential worlds \textbf{H} expects to see. 

$\mathit{K}$ is considered to assess its impact on the planner's performance. 
We assume that whenever the shared execution experience is disrupted, 
\textbf{R} can execute a \textit{maximum} of $\mathit{K}$ actions, including the option of doing nothing. 
For example, when the human is away to fetch the cube and has a \textit{fixed} length and sequence of actions to perform.  
The exact number of real ontic actions \textbf{R} performs ranging from $0$ to $\mathit{K}$, including which of those allowed ones and their potential sequences, will depend on the scenario at hand, environment dynamics (e.g., the observability factor), and the optimization criteria.  
The option for the robot to limit its real actions whenever required is integrated into the task description, aligning with the turn-taking nature of the underlying planner.
Consequently, the planner is engineered to optimize the robot's policy tree branching on uncontrollable human choices, including a communication action, to meet our objective.

%
\paragraph{\em \textbf{Qualitative Analysis.}}
In our use case domain, we explore different plan traces the planner can come up with depending on scenarios that arise.
We start with two cubes, $c_r$ and $c_w$, placed initially on tables $mt$ and $ot$, respectively. 
Initially, there is only one designated world, $w_d$, in the initial epistemic state, $s_0$. 
The environment otherwise remains unchanged. 
\textbf{H} can decide to go and retrieve the white cube, while the robot begins to work on other parts of the shared task. 

\begin{figure}[!t]
    \centering    \includegraphics[width=0.54\textwidth]{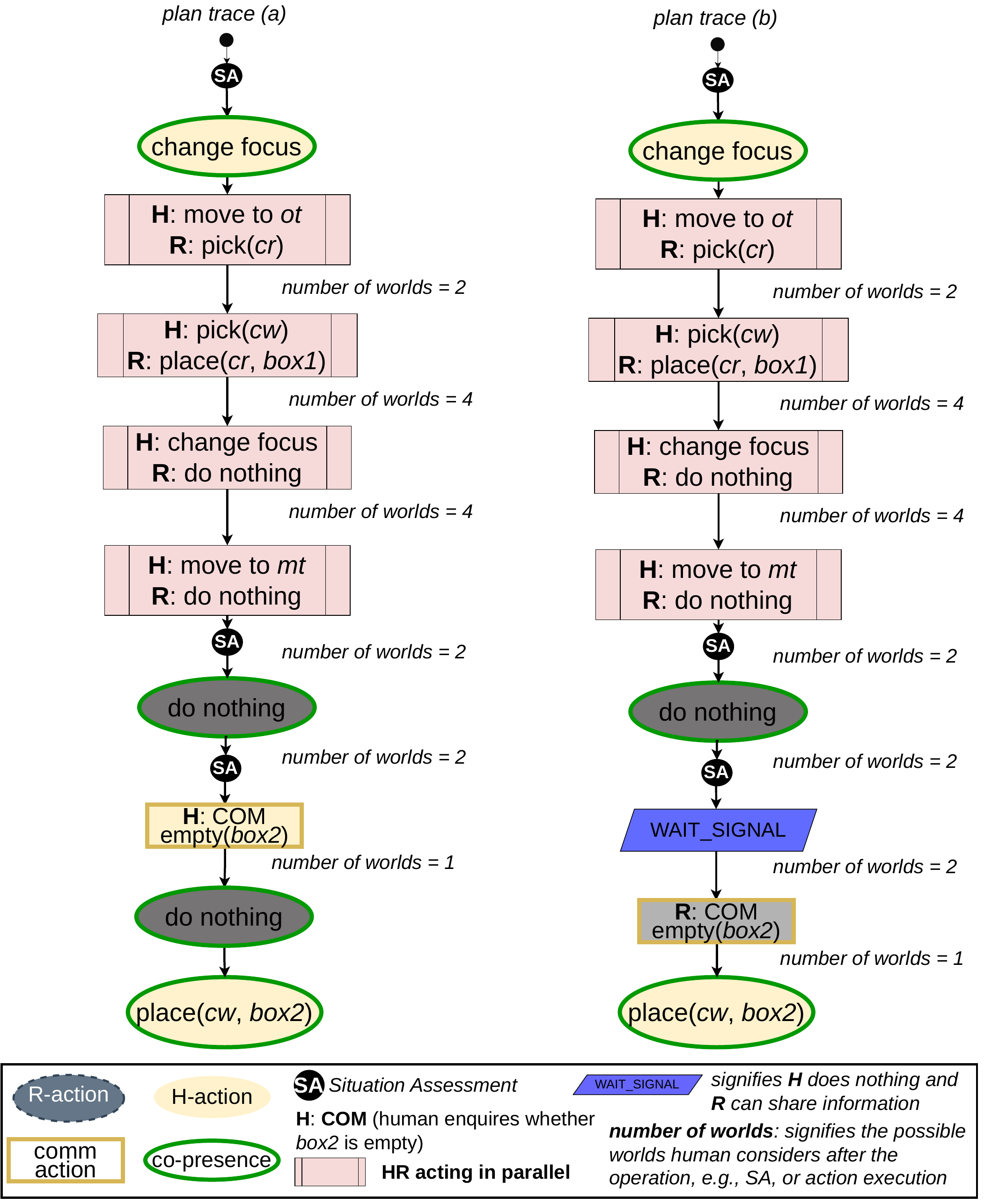}
    \caption{
    {\em Two branches from an AND/OR joint solution are shown: (a) \textbf{R} informs \textbf{H} {\em proactively}, thus leaving only the designated world for them to continue with {\em place($c_w$, $box_2$)}. (b) \textbf{R} {\em waits} to inform \textbf{H} about the condition {\em empty($box_2$)}. 
    }}
    \label{fig:qualitative-plans}
    \vspace{-0.6cm}
\end{figure}

Two plan traces are shown in Figure~\ref{fig:qualitative-plans}. 
\textbf{H} starts to execute. \textbf{H}\&\textbf{R} are co-present and the boxes are opaque. 
(\textit{SA} is shown only at relevant places.) 

Let us focus on (a): after the human shifts focus to $ot$, both agents are not co-present until they reunite later in the trace, during which they act simultaneously.  
(\textit{In this situation, agents must be at the same table and simultaneously focus on it to be considered co-present.}) 
In the first broad rectangular box, the human moves to $ot$. They anticipate that the robot may have picked $c_r$ or done nothing, but in reality, the robot picks $c_r$, resulting in two possibilities that will be maintained within the robot.
Similarly, in the following box, the human picks $c_w$ at $ot$ and anticipates that if the robot had picked $c_r$, it could have placed it in one of the boxes or held onto it, or $c_r$ is still on the table. 
Together, these create four possibilities, with the reality being that $c_r$ is inside $box_1$. 
At this point, the robot currently has no feasible action to execute, and the shared task has been not achieved yet, too. 
Upon the human's return, as per their initial agreement on $\mathit{K}$, the robot has prepared itself with four possible worlds (with a designated world that only the robot knows). 
Perspective-taking and situation assessment help the robot eliminate two worlds where $c_r$ is not on $mt$ or in \textbf{R}'s hand.


We present two approaches to proceed with the task. 
In trace (a), the robot waits for human inquiry, while in trace (b), the human does nothing.  
Consequently, the robot decides to inform that $box_2$ is empty, resulting in only the designated world remaining. Here, $empty(box_2)$ is a precondition for the human to place $c_w$ in it, which is true in one world and not another. 
Our proposed method considers a situation where the human waits for information without taking any action, such as nodding or making eye contact with the robot, as a distinct condition (\textit{trace (b)}). 
Additionally, \textbf{R} can signal the value of \textit{p} to \textbf{H} by manipulating a variable \textit{q} (inline with ~\cite{shekhar2021improved}), which we aim to explore further.

In the 3-cube scenario, if $c_r$ is already in $box_1$ and \textbf{R} is holding $c_y$, it can choose to place the $c_y$ in $box_1$ in the presence of \textbf{H}. 
This action results in the creation of a state with only the designated world as the next action ordered in the task network ($tn_{r_h}$) of that world does not allow \textbf{R} to execute \textit{place($c_y, box_1)$}. 
The robot can only be clever if it can fully explore its options. Depending on the situation, it might not always be preferable to place the yellow cube while the human is away and rely on communication or other means later on.


In contrast, in~\cite{FavierSA23}, \textbf{R} communicates immediately after agents reunite.
This assumes that \textbf{H} can choose to place $c_w$ in $box_1$ due to their outdated belief. In some practical cases, not communicating may lead to detrimental effects.



\paragraph{\em \textbf{Quantitative Results and Analysis.}}
%
Refer to Table~\ref{tab:quant_res_1}. 
In each instance, at least one cube is positioned on $ot$, which \textbf{H} must retrieve. 
We show how the factor $\mathit{K}$ influences the overall runtime. 

We observe that $|W|$ and $\mathit{K}$ contribute to longer runtime in both domains. Instances requiring communication tend to take slightly longer compared to those where communication is not required.










\section{Conclusion }
Our framework allows the robot to implement a ToM not only at execution time but also at planning time and hence explores what would be the beliefs of the human and the robot depending on which course of action. 
This is done thanks to the use of epistemic reasoning, the notion of shared experience, and observable and non-observable facts, which allow anticipation of \textbf{H}'s situation assessment along the various non-deterministic shared plan traces of \textbf{H} and \textbf{R}.


\textbf{R} can adapt its choices to \textbf{H}'s diverging beliefs over time, e.g. by choosing to communicate to inform \textbf{H} or elicit an action, or a particular context to act.

We acknowledge that scaling such abilities can pose complexity challenges for planners, which can be evident in~\cite{KR2021-12}.
Hence, we take care to precisely identify the context in which our approach can be effectively used which is dealing in a refined manner with
short-term interactions and intricate \textbf{H}\&\textbf{R} face-to-face situations.
Also, we intend to test the current system in different domains with realistic \textbf{H}\&\textbf{R} co-activities. 
We aim to enhance planner's practical efficiency and explore incremental task planning.      

%

\noindent 


\noindent \textbf{User Study:}
We tested with users the HATP framework, which supports execution concurrency and demonstrated the robot's ability to adapt to non-deterministic human behaviors~\cite{Favier-thesis-24,favier-acs-24}. Although this study is not for testing advanced epistemic reasoning of EHATP, it offers valuable insights and tools. 

Building on these findings, we are evaluating the EHATP framework, which incorporates features such as second-order theory of mind and belief divergence. 

\vspace{-0.1in}
{
\paragraph{\em \textbf{Acknowlegments:}} This work has been partially funded by the Agence Nationale de la Recherche through the ANITI ANR-19- PI3A-0004 grant and the Horizon Europe Framework Programme through the euROBIN Grant 101070596.}





%
%
\bibliographystyle{splncs04}
\bibliography{mybibliography}
%





\end{document}